\documentclass{article}

 \usepackage[preprint]{neurips_2023}

\usepackage[utf8]{inputenc} 
\usepackage[T1]{fontenc}    
\usepackage{hyperref}       
\usepackage{url}            
\usepackage{booktabs}       
\usepackage{amsfonts}       
\usepackage{nicefrac}       
\usepackage{microtype}      
\usepackage{xcolor}         
\usepackage{algorithm}
\usepackage{algpseudocode}
\usepackage{mathtools}
\usepackage{array}
\usepackage{verbatim}
\usepackage{float}

\usepackage{graphicx}
\graphicspath{{figures/}}

\title{Projection-Free CNN Pruning via Frank-Wolfe with Momentum: Sparser Models with Less Pretraining}

\author{%
\textbf{Hamza ElMokhtar Shili}$^{1}$ \quad Natasha Patnaik$^{2}$ \quad Isabelle Ruble$^{2}$ \\ \textbf{Kathryn Jarjoura}$^{2}$ \quad \textbf{Daniel Suarez Aguirre}$^{2}$ \\ \\
$^1$Rice University, Computer Science Department \\ \\
$^2$Rice University, Computational Applied Math \& Operations Research Department \\ 
}

\begin{document}

\maketitle

\begin{abstract}
We investigate algorithmic variants of the Frank–Wolfe (FW) optimization method for pruning convolutional neural networks. This is motivated by the ``Lottery Ticket Hypothesis'', which suggests the existence of smaller sub-networks within larger pre-trained networks that perform comparatively well (if not better). Whilst most literature in this area focuses on Deep Neural Networks more generally, we specifically consider Convolutional Neural Networks for image classification tasks. Building on the hypothesis, we compare simple magnitude-based pruning, a Frank–Wolfe style pruning scheme, and an FW method with momentum on a CNN trained on MNIST. Our experiments track test accuracy, loss, sparsity, and inference time as we vary the dense pre-training budget from 1 to 10 epochs. We find that FW with momentum yields pruned networks that are both sparser and more accurate than the original dense model and the simple pruning baselines, while incurring minimal inference-time overhead in our implementation. Moreover, FW with momentum reaches these accuracies after only a few epochs of pre-training, indicating that full pre-training of the dense model is not required in this setting.

\end{abstract}

\section{Problem Statement and Notation}

Convolutional neural networks are ideal models for image classification and related computer vision tasks. However, large models with many parameters pose serious challenges - both in terms of training time and memory. In this paper, we motivate and outline a series of computational experiments aimed at evaluating different variants of the Frank-Wolfe (FW) algorithm for neural network pruning, intending to find well-performing sparser models.

\paragraph{Notation - From Deep Neural Networks to Convolutional Neural Networks.}

To represent an arbitrary neural network with $N$ neurons, we follow the notation used in \cite{wolfe2023pretraining} and \cite{ye2020goodsubnetworks}. Let $\mathbf{x} \in \mathbb{R}^{d}$ denote the input vector and $\mathbf{\theta_{i}}$ for $i \in [N] = \{1,..., N\}$ denote the parameters/ weights associated with the i-th neuron. With these inputs, the i-th neuron can be represented as a function $\sigma(\mathbf{x}, \mathbf{\theta_{i}})$. 

Although this function $\sigma(\mathbf{\cdot}, \theta_{i})$ can take on general forms, \cite{wolfe2023pretraining} and \cite{ye2020goodsubnetworks} consider the specific case of a two-layer network for simpler analysis (a representative example of more complex architectures). In the two-layer case, we can write $\sigma(\mathbf{x}, \mathbf{\theta_{i}}) = b_i \cdot \sigma_{+}(\mathbf{a_i}^{T}\mathbf{x})$, where $\mathbf{\theta_{i}} = [b_i \ \ \mathbf{a_i}]$ denotes the concatenation of the second layer weight, $b_i$, and the vector of first layer weights, $\mathbf{a_i} \in \mathbf{R}^{d}$, respectively. The function $\sigma_{+}(\cdot)$ represents the activation function. 

Accordingly, we use the following notation for the output of a neural network containing $N$ neurons: 
\begin{align}\label{denseloss}
    f_{[N]}(\mathbf{x}, \Theta) = \frac{1}{N} \sum_{i=1}^{N} \sigma(\mathbf{x}; \mathbf{\theta_{i}})
\end{align}

Given a dataset $D = \{\mathbf{x}^{(j)}, y^{(j)}\}_{i=1}^{m}$ consisting of $m$ observations, each with attributes $\mathbf{x}^{(j)} \in \mathbb{R}^{d}$ and desired output $y^{(j)} \in \mathbb{R}$, we can train the neural network to discover the set of weights $\Theta$ that minimizes the following loss function:
\begin{align}
    \mathcal{L}\bigl[f_{[N]}(\cdot, \Theta)\bigr] = \frac{1}{2} \sum_{j=1}^{m} \bigl( f_{[N]}(\mathbf{x}^{(j)}, \Theta) - y^{(j)} \bigr)^{2}
\end{align}

While existing analysis in this area uses the representative example of a two-layer network, Convolutional Neural Networks have a more complex architecture, which includes convolutional layers and pooling layers. To extract local features, convolutional layers apply filters that learn to detect specific patterns. The output of a convolutional layer $l$, denoted $\mathbf{h}^{(l)}$, is obtained by convolving the input feature maps $\mathbf{h}^{(l-1)}$ with learnable filters $\mathbf{W}^{(l)}$ and applying an activation function, $\psi$, that introduces non-linearity (such as ReLU):
\begin{equation}
\mathbf{h}^{(l)} = \psi\bigl(\mathbf{W}^{(l)} * \mathbf{h}^{(l-1)}\bigr)
\end{equation}

Such convolutional layers are typically succeeded by pooling layers, which reduce spatial dimensions of the feature maps to create a more compact representation while retaining essential information. For instance, max pooling will simply select the maximum value within a given local region:
\begin{equation}
\mathbf{h}_{\text{pool}}^{(l)} = \text{Max}\bigl\{\mathbf{h}^{(l)}\bigr\}
\end{equation}

\paragraph{Three Stage Approach for Neural Network Sparsification} The classic Stochastic Gradient Descent (SGD) algorithm is commonly used for Neural Network training, with the model's weights $\Theta = \{\theta_{1}, ..., \theta_{N}\}$ typically being unconstrained \cite{pokutta2020deep}. As outlined in \cite{liu2019}, \cite{wolfe2023pretraining}, and \cite{ye2020goodsubnetworks}, the computational cost and memory requirements for training large-scale, dense models as $N$ increases necessitates a more practical three-step procedure for finding smaller sub-networks: \textit{(i) Pretraining} - minimize \eqref{denseloss} by performing SGD over $\Theta$, \textit{(ii) Pruning} - finding a sub-network from the original model which performs well, and \textit{(iii) Retraining} - performing SGD again to maintain desired accuracy, either using the newer model's weights as the initial parameters (as in \cite{ye2020goodsubnetworks}) or starting with a new random parametrization (as in \cite{liu2019}).

In particular, the pruning stage is concerned with finding a subset of neurons $S \subset [N]$ which minimizes the loss of the newly defined sub-network: $f_{S}(\mathbf{x}, \Theta) = \nicefrac{1}{|S|} \sum_{i\in S} \sigma(\mathbf{x}; \mathbf{\theta_{i}})$. In particular, we would like this sub-network's accuracy to be comparable to the original dense model:
\begin{align}
     \mathcal{L}\bigl[f_{[S]}\bigr] \geq \mathcal{L}\bigl[f_{[N]}\bigr] - \delta \text{ for some } \delta \in \mathbb{R}^{+}
\end{align}

This allows us to find sub-networks that perform relatively well (as compared to the original dense model) \cite{wolfe2023pretraining}. Furthermore, in \cite{ye2020goodsubnetworks}, they find that pruning dense models guarantees finding more accurate sparse networks than directly training smaller models can, which is in line with the Lottery Ticket Hypothesis \cite{lotteryticket}.

\section{Background}

\paragraph{Lottery Ticket Hypothesis} The Lottery Ticket Hypothesis (LTH), proposed by Frankle and Carbin in 2019, suggests that within randomly-initialized, dense neural networks, there exist sparse subnetworks, or "winning tickets," capable of achieving test accuracy comparable to the original network in a similar number of iterations. These winning tickets benefit from fortuitous initializations that facilitate effective training. Justified by the LTH, a structured three-stage approach to neural network sparsification emerges.\\\\
Firstly, in the identification stage, pruning techniques are employed to systematically remove connections from the network, thereby identifying these winning tickets. Pruning enables significant reductions in model size by removing unnecessary parameters while preserving the network's essential functionality. Frankle and Carbin's reserach consistently found that winning tickets are less than $10-20\%$ of the size of several fully-connected and convolutional architectures for MNIST and CIFAR10 datasets \cite{lotteryticket}.\\\\
Secondly, in the optimization stage, winning tickets are leveraged to design more efficient training schemes. By focusing on training these sparse subnetworks from the start, the aim is to accelerate convergence and improve overall training dynamics. This stage involves devising novel training algorithms and optimization strategies tailored to the unique properties of winning tickets, enabling faster learning and improved generalization.\\\\
Lastly, in the refinement stage, iterative pruning techniques are employed to further enhance the performance of the identified winning tickets. Iterative pruning outperforms one-shot pruning in identifying winning tickets by repeatedly training, pruning, and refining the network over multiple rounds. This iterative process results in networks that learn faster and achieve higher test accuracy, ultimately leading to more efficient and effective deep learning models \cite{lotteryticket}.\\\\
These implications suggest that certain sparse architectures with fortuitous initializations possess inherently better training properties and offer opportunities for efficient compression, which involves reducing the size of neural networks while maintaining performance. This structured approach not only advances the field of neural network pruning and optimization but also lays the foundation for more efficient and effective deep learning models.

\paragraph{Suitable Feasible Regions} 
Frank-Wolfe methods can offer computational efficiency by minimizing a linear objective over the feasible set without needing costly projections \cite{wang2021frankwolfe}. This advantage extends to scenarios with polyhedral and nuclear norm ball constraints, simplifying the linear subproblems and streamlining solutions. Additionally, their projection-free nature makes them popular, particularly in machine learning, where they surpass methods requiring projection at each iteration. As such, Frank-Wolfe methods are appropriate when the optimization problem's constraints define convex, differentiable, and compact feasible regions\cite{jaggi13}. Convexity ensures that the feasible region forms a convex set, enabling efficient exploration of the solution space. Differentiability allows for the use of gradient information to guide the optimization process towards the optimal solution. Compactness ensures that the feasible region is bounded, facilitating convergence towards a global optimum.\\\\
These essential characteristics of feasible regions crucial for the effective implementation of Frank-Wolfe algorithms, particularly within the realm of large-scale optimization, has been studied as the effectiveness of the algorithm was shown to be intricately tied to the convexity and compactness of the feasible region, coupled with the ability to solve linear optimization subproblems efficiently within this region \cite{ding2018frankwolfe}. Specifically, Frank-Wolfe methods excel in scenarios where constraints define bounded convex regions, enabling swift exploration of the solution space without necessitating costly projection operations. This alignment fits the needs of neural network pruning tasks, where a loss function subject to constraints on network sparsity is minimized. The convexity of the feasible region also ensures a well-defined optimization problem, an essential factor in steering the iterative pruning process efficiently towards an optimal solution. Additionally, the compactness property guarantees convergence towards a global optimum, ensuring the algorithm's resilience in traversing the solution landscape. Furthermore, the differentiability of the feasible region enables the computation of gradients, essential for guiding the pruning process towards optimal sparse architectures efficiently. Compactness ensures a bounded search space, preventing the optimization process from straying towards infeasible solutions.\\\\
By satisfying these properties, the optimization problem in neural network pruning mirrors the characteristics of problems well-suited for Frank-Wolfe methods. The iterative nature of Frank-Wolfe-style approaches, where solutions are iteratively refined based on linear approximations of the objective function within the feasible region, aligns with the iterative pruning process. Moreover, the reliance on gradients to navigate the search for optimal sparse architectures underscores the significance of differentiability, a core property of Frank-Wolfe methods.

\section{Convolutional Neural Network Pre-training and Pruning}

\paragraph{Convolutional Neural Networks}
Convolutional Neural Networks (CNNs) are a class of neural networks primarily utilized for image classification. The defining feature of CNNs is the convolution operation, which involves sliding filters over the input data to extract and learn important features such as edges and textures. Key elements of CNNs include:

\begin{itemize}
    \item \textbf{Convolutional Layers}: Detect features by applying various filters to the input.
    \item \textbf{Activation Functions}: Introduce non-linearity into the model, enabling it to learn complex patterns. Eg. ReLU
    \item \textbf{Pooling Layers}: Reduce the dimensionality of each feature map while retaining the most important information.
    \item \textbf{Fully Connected Layers}: After the convolutional and pooling layers extract and reduce features, these layers classify the input based on the detected features by outputting probabilities over the classes.
\end{itemize}

\paragraph{Resource Constrained Neural Networks and Pruning Paradigm}
While machine learning and deep learning applications are increasing at an exorbitant rate, their real-world implementation is delayed by their high computational requirements in the form of cutting-edge hardware, large memory, and high energy use. \cite{chin2018layercompensated}, \cite{Pietrołaj_Blok_2024} With this context, resource-constrained solutions arise. Henceforth, we will show how pruning can help to improve efficiency in training and inference. 

Neural network pruning consists of simplifying a neural network by setting some weights to zero according to some rule. For example, a basic pruning rule may consist of removing weights smaller than a certain threshold. Overall, the goal of running is to reduce model size and complexity without compromising performance. In general, the pruning process is the following:

\begin{itemize}
    \item \textbf{Pre-Training}: Train NN to get a baseline model.
    \item \textbf{Pruning}: Introduce non-linearity into the model, enabling it to learn complex patterns.
    \item \textbf{Re-Training}: Reduce the dimensionality of each feature map while retaining the most important information.
\end{itemize}

Note that pruning and re-training may be done in an iterable manner; this is known as the tunning stage. Also, under some methods, re-training may not be necessary given the needs of the user

Within the perspective of resource-constrained neural networks, a pruned network is smaller than the baseline model and this achieves better efficiency for inference tasks. From the angle of training, pruning may also be beneficial because rather than training a large network to a full extent, the large network is trained only up to a point from which a sufficiently good subnetwork is identified. Therefore, even when considering pre-training and re-training, pruning may lead to using fewer resources as, after pruning, the rest of training is done in a smaller network. 

Finally, note that this framework also applies in the case of CNNs, which are the base of our computational study.

\paragraph{Frank-Wolfe for Pruning} To find well-performing smaller sub-networks within dense models that are pre-trained via stochastic GD, Frank-Wolfe-style approaches can be used \cite{wolfe2023pretraining} \cite{ye2020goodsubnetworks}, where the desired sparsity level is treated as a constraint. Thus, even though Neural Network training itself is a non-convex optimization problem, the pruning stage can be reformulated as a constrained convex optimization problem, with Frank-Wolfe-style approaches being used \cite{wolfe2023pretraining}. 

\paragraph{Frank-Wolfe in Machine Learning Context} The classic FW algorithm (and related projection-free methods) provides an alternative to projected Gradient Descent for convex-constrained optimization problems. Its low
per-iteration complexity and suitability for complicated constraints makes it very effective in the context of large-scale machine-learning problems \cite{pokutta2024intro}.

FW replaces the projection step of projected GD with a linear minimization sub-problem that can be described as follows:

We form the next iterate by taking a convex combination between (i) the previous iterate and (ii) an extreme point of the feasible region that best approximates the gradient direction. \cite{pokutta2024intro} This ensures we retain feasibility within each iteration, without using a projection step. Thus, FW is particularly advantageous when the traditional projection step in projected GD is computationally expensive \cite{jaggi13}. See the pseudocode description for the classic deterministic variant below:

\begin{algorithm}[H]
\caption{Classic (Deterministic) Frank-Wolfe Algorithm \cite{fw1956original}, \cite{jaggi13}}\label{alg:FW}
\begin{algorithmic}
\State \textbf{Input: } $x_0 \in \mathcal{C} \coloneq$ initial starting point; $T \coloneq$ number of iterations
\vspace{0.2em}
\For{$t = 0, 1, ..., T$}
\vspace{0.2em}
    \State $v_t \gets \text{argmin}_{x_t \in \mathcal{C}} \langle x_t, \nabla f(x_t)\rangle$ \Comment{Approximate gradient direction, maintaining feasibility}
    \vspace{0.2em} 
    \State $\gamma \gets  \frac{2}{t+2}$ 
    \vspace{0.2em} 
    \State $x_{t+1} \gets (1-\gamma)x_t + \gamma v_t$ 
    \Comment{Convex combination: previous iterate and $v_t$}
    \vspace{0.2em}
\EndFor
\end{algorithmic}
\end{algorithm}

In Algorithm \eqref{alg:FW}, we can also substitute the linear minimization $v_t = \text{argmin}_{v_t \in \mathcal{C}} \langle x_t, \nabla f(x_t)\rangle$ with an approximate approach, as opposed to solving for the minimizer exactly \cite{jaggi13}. Another simple modification is to perform a line search within each iteration $t$ for the optimal $\gamma$ required to find the best next iterate on the line segment connecting $x_t$ and $v_t$ \cite{jaggi13}. 

\section{Pruning Algorithms}

\paragraph{Simple Pruning}
The first algorithm takes a naive approach to pruning. First, all of the weights of the model are retrieved and sorted based on magnitude. A certain percentage of the weights with the lowest magnitude are set to zero based on a user-defined \textit{pruning percentage}.

\begin{algorithm}[H]
\caption{Simple Pruning}
\label{alg:ModelPruning}
\begin{algorithmic}
\State \textbf{Input:} $M \in \mathcal{M}$, a neural network model; $p \in [0,1]$, pruning percentage; $D$, training dataset; $E$, epochs for retraining
\vspace{0.2em}
\State $W \gets$ weights of $M$ \Comment{Retrieve model weights}
\vspace{0.2em}
\State $n \gets |W|$ \Comment{Total number of weights}
\vspace{0.2em}
\State $n_p \gets \text{round}(p \cdot n)$ \Comment{Compute number of weights to prune}
\vspace{0.2em}
\State $W_f \gets$ flattened $W$ \Comment{Concatenate all weights into a vector}
\vspace{0.2em}
\State $\text{idxs} \gets$ indices sorted by $|W_f|$
\vspace{0.2em}
\State $I_p \gets \text{first } n_p \text{ elements of } \text{idxs}$
\vspace{0.2em}
\For{$i \in I_p$}
    \State $W_f[i] \gets 0$ \Comment{Set pruned weights to zero}
    \vspace{0.2em}
\EndFor
\State Update $W$ of $M$ with $W_f$
\vspace{0.2em}
\State Set optimizer, loss, and metrics for $M$
\vspace{0.2em}
\State Retrain $M$ on $D$ for $E$ epochs
\vspace{0.2em}
\State \textbf{return} $M$
\end{algorithmic}
\end{algorithm}

\paragraph{Frank-Wolfe Pruning}
The second algorithm considers a randomly chosen subsample of the weights and takes gradient information into account when selecting weights to prune. Rather than pruning a fixed percentage of the weights, the pruning mask is applied  based on a user-defined \textit{target sparsity level}. This method takes into account cases in which smaller weights are actually more integral to the structure of the CNN and thus is more flexible than the simple pruning algorithm.

\begin{algorithm}[H]
\caption{Frank-Wolfe Pruning}
\label{alg:FrankWolfePruning}
\begin{algorithmic}
\State \textbf{Input:} $M \in \mathcal{M}$, neural network model; $D$, training dataset; $s$, target sparsity level; $N$, number of pruning iterations; $S$, subsample size; $E$, epochs for retraining
\vspace{0.2em}
\State Initialize optimizer (SGD)
\vspace{0.2em}
\For{$i = 1$ to $N$}
    \State $G \gets$ zeros (shape of $M$'s trainable weights)
    \State $subsample \gets$ sample $S$ instances from $D$
    \For{each $(x, y)$ in $subsample$}
        \State $G \gets G + \nabla \text{loss}(M, x, y, \text{optimizer})$
    \EndFor
    \If{$i \mod \left(\frac{N}{2}\right) = 0$}
        \State Apply pruning to $M$ based on $G$ and sparsity $s$
    \EndIf
    \State Retrain $M$ using $D$ for $E$ epochs
\EndFor
\State \textbf{return} $M$
\end{algorithmic}
\end{algorithm}

\paragraph{Frank-Wolfe Pruning + Momentum}
The third algorithm has two main modifications compared to the basic frank-wolfe method: dynamic sparsity and momentum. The \textit{target sparsity level} is determined by the user. However, the sparsity of the model is increased iteratively which potentially minimizes large decreases in model performance at each iteration. Additionally, a user-defined \textit{momentum} vector is applied to the gradients in order to further differentiate the weights and stabilize the algorithm.

\begin{algorithm}[H]
\caption{Frank-Wolfe Pruning + Momentum}
\label{alg:IterativePruning}
\begin{algorithmic}
\State \textbf{Input:} $M \in \mathcal{M}$, neural network model; $D$, training dataset; $s_{\text{init}}$, initial sparsity; $s_{\text{final}}$, final sparsity; $m$, momentum parameter; $N$, number of iterations; $E$, fine-tuning epochs; $S$, subsample size; Optimizer
\vspace{0.2em}
\State $\Delta s \gets \frac{s_{\text{final}} - s_{\text{init}}}{N}$
\State $s \gets s_{\text{init}}$
\vspace{0.2em}
\For{$i = 1$ to $N$}
    \State $G \gets$ zeros (shape of $M$'s trainable weights)
    \State $subsample \gets$ sample $S$ instances from $D$
    \For{each $(x, y)$ in $subsample$}
        \State $G \gets G + \nabla \text{loss}(M, x, y, \text{Optimizer})$
    \EndFor
    \State $G \gets m \cdot G$ \Comment{Apply momentum to gradients}
    \State Apply pruning to $M$ based on $G$ and sparsity $s$
    \If{$E > 0$}
        \State Retrain $M$ using $D$ for $E$ epochs
    \EndIf
    \State $s \gets \min(s + \Delta s, s_{\text{final}})$
\EndFor
\State \textbf{return} $M$
\end{algorithmic}
\end{algorithm}

\section{Numerical Experiments}

\paragraph{Datasets} For our computational study, we consider image classification problems. Here we used the MNIST \cite{lecun-mnisthandwrittendigit-2010} dataset as our baseline. This dataset contains handwritten digits, with 60,000 examples in the training set and 10,000 examples on the test set. The digits are centered in 28x28 grayscale images. The only pre-processing applied to the data is normalization of the grayscale by dividing the value of each pixel by 255, ensuring that the pixels used as input for the model are between 0 and 1.

\paragraph{Network Architecture}
For our numerical analysis, we employ a CNN designed for the classification of 28x28 images into 10 categories, as is the case for the MNIST dataset. The architecture is detailed in Table 1.

\begin{table}[h!]
\centering
\begin{tabular}{|>{\raggedright\arraybackslash}p{3cm}|>{\raggedright\arraybackslash}p{5.5cm}|>{\raggedright\arraybackslash}p{5.5cm}|}
\hline
\textbf{Layer Type} & \textbf{Configuration} & \textbf{Purpose} \\ \hline
Convolutional & 32 filters of size 3x3, ReLU activation & Captures basic features from input images, introduces non-linearity. \\ \hline
Max Pooling & Pool size of 2x2 & Reduces spatial dimensions, making the detection of features somewhat invariant to scale and orientation changes. \\ \hline
Convolutional & 64 filters of size 3x3, ReLU activation & Increases the complexity of the model to capture more detailed features. \\ \hline
Max Pooling & Pool size of 2x2 & Further reduces spatial dimensions, focusing on the most important features. \\ \hline
Flatten & --- & Transforms 3D feature maps into 1D feature vectors. \\ \hline
Dropout & Dropout rate of 0.5 & Prevents overfitting by randomly setting input units to 0 during training. \\ \hline
Dense & 10 units, Softmax activation & Outputs the probability distribution over the ten classes. \\ \hline
\end{tabular}
\caption{CNN Architecture for MNIST Classification}
\label{tab:architecture}
\end{table}

\paragraph{Training and pruning setup.}
We train the base CNN on MNIST using stochastic gradient descent and sparse categorical cross-entropy. We vary the dense pre-training budget from 1 to 10 epochs; for each budget we save a checkpoint of the dense model and then apply each pruning strategy independently to that checkpoint. All pruning methods act only on the convolutional and fully connected layers; pooling, flatten, and dropout layers are left unchanged. The simple pruning baseline removes weights using a greedy backward-selection rule until a target sparsity is reached, followed by a short fine-tuning phase. The Frank–Wolfe (FW) variants use iterative pruning: at each iteration we estimate weight importance on a small subsample of the training data, prune in the FW direction, and fine-tune for a few epochs. In the FW+momentum variant, the FW direction is replaced by an exponential moving average of past gradients, so sparsity increases gradually from an initial value $s_{\text{init}}$ to a final target $s_{\text{final}}$. In all subsequent figures, the horizontal axis (“Epochs”) should therefore be read as \emph{pre-training budget}: each point corresponds to a different dense checkpoint pruned and fine-tuned independently, not to continued training of a single model.

\paragraph{Results and Analysis}

\begin{figure}[t]
    \centering
    \includegraphics[width=0.8\textwidth]{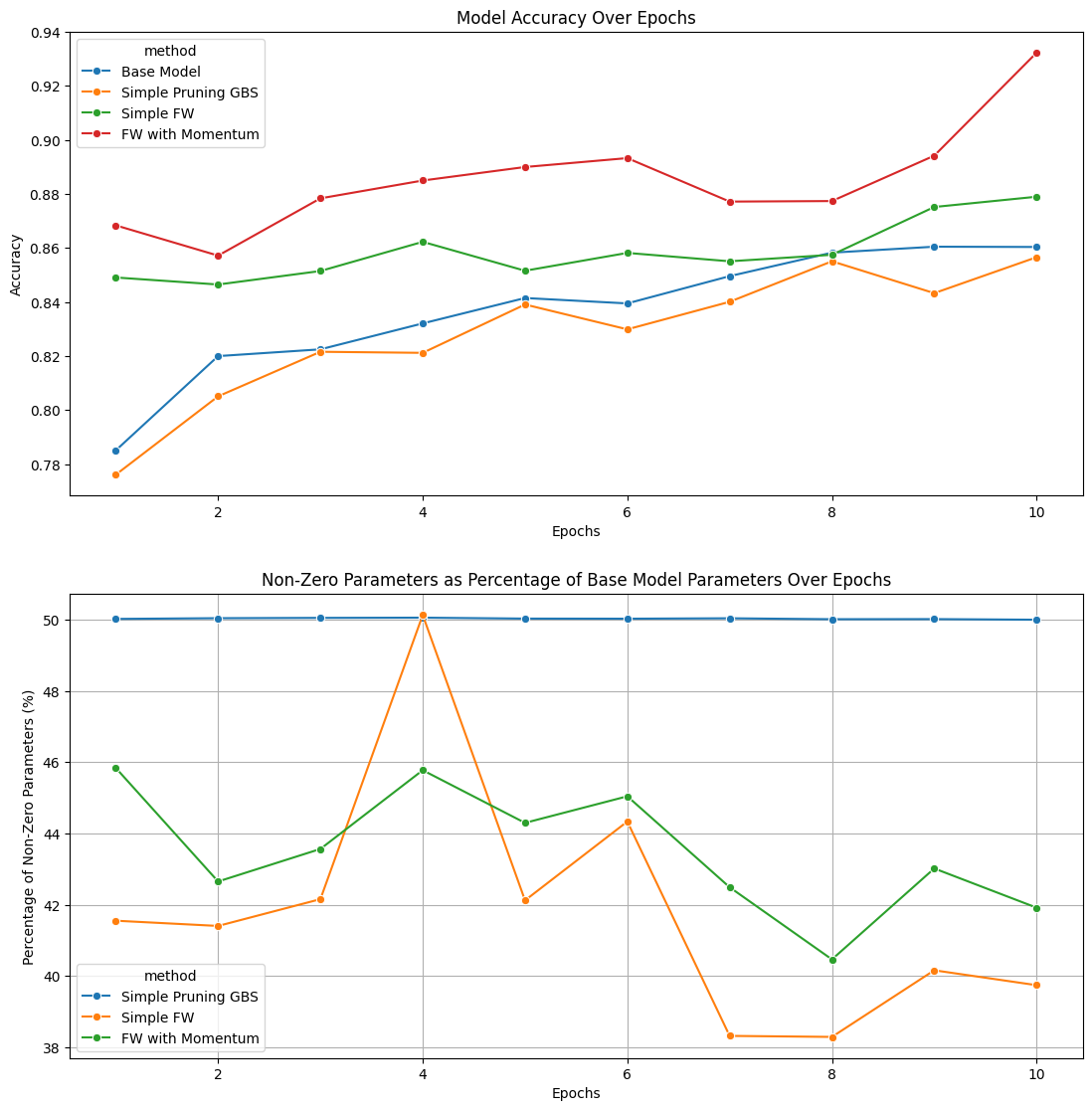} 
    \caption{Test accuracy (top) and loss (bottom) of pruned and unpruned models as a function of pre-training budget (epochs).}
    \label{fig:accuracy_loss}
\end{figure}

\begin{figure}[t]
    \centering
    \includegraphics[width=0.8\textwidth]{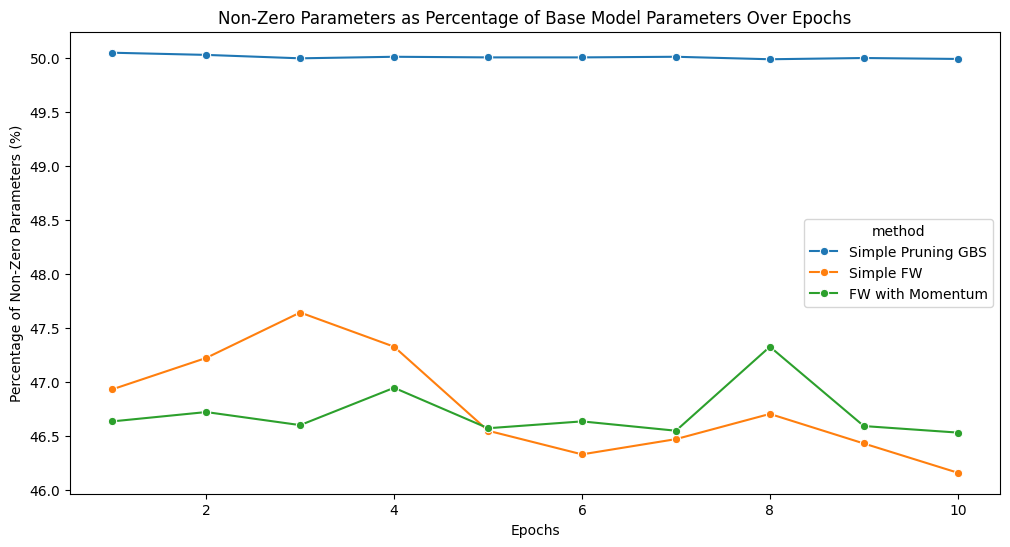}
    \caption{Percentage of non-zero parameters in neural networks post-pruning, relative to the unpruned base model, across different pre-training budgets (epochs).}
    \label{fig:non_zero_parameters}
\end{figure}

\begin{figure}[t]
    \centering
    \includegraphics[width=0.8\textwidth]{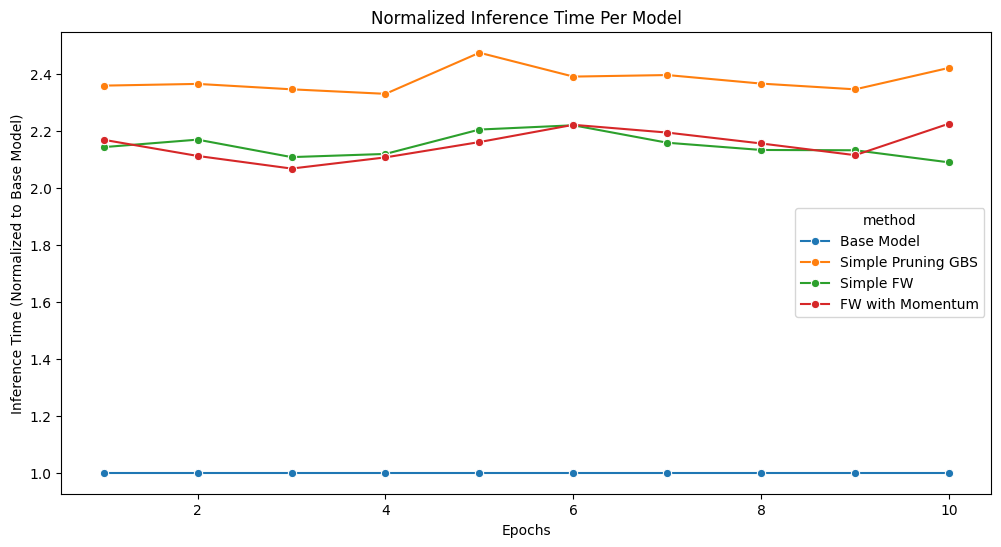}
    \caption{Normalized inference time on the test set (lower is better) for each pruning method, as a function of pre-training budget (epochs).}
    \label{fig:inference_time}
\end{figure}

We evaluate the pruning methods in terms of test accuracy, test loss, normalized inference time, and the fraction of non-zero weights, as shown in Figures~\ref{fig:accuracy_loss}–\ref{fig:inference_time}.

\textbf{Accuracy and loss.}
Figure~\ref{fig:accuracy_loss} (top) tracks test accuracy as we vary the dense pre-training budget from 1 to 10 epochs. All methods improve with more pre-training, but the ordering between pruning strategies is consistent. With just a single pre-training epoch, the base model and simple pruning reach only about $0.78$–$0.79$ test accuracy, whereas FW and FW+momentum are already in the mid-$0.80$s. After 10 epochs, FW+momentum reaches roughly $0.94$ accuracy, compared to about $0.88$ for FW, $0.85$ for simple pruning, and a similar range for the unpruned baseline.

Interpreting the x-axis as pre-training budget makes the efficiency gain clear. After only \emph{1–2 pre-training epochs}, FW+momentum already reaches about $0.86$–$0.88$ test accuracy, matching or exceeding the accuracy of the dense baseline and simple pruning baselines even when they are pre-trained for 8–10 epochs (approximately $0.85$–$0.86$). In other words, FW+momentum lets us stop dense pre-training much earlier while still attaining, and often improving on, the final accuracy of models that were fully pre-trained.

The loss curves in Figure~\ref{fig:accuracy_loss} (bottom) exhibit the same pattern. FW+momentum yields the lowest test loss at every pre-training budget, ending around $0.28$ after 10 epochs, substantially below both the base model and simple pruning, and noticeably below FW without momentum. Simple pruning, in contrast, never clearly improves on the base network and can slightly worsen both accuracy and loss when pre-training is short.

\textbf{Model complexity.}
Figure~\ref{fig:non_zero_parameters} reports the percentage of non-zero parameters remaining after pruning, relative to the base model. The simple pruning baseline hovers close to $50\%$ of the original weights across all pre-training budgets. Both FW variants produce consistently sparser networks: they remove several additional percentage points of weights beyond the simple method, with FW+momentum always at or below the sparsity achieved by vanilla FW. Even though the absolute gap is modest in this small CNN, it shows that FW+momentum is more aggressive at removing redundant filters while still improving accuracy.

\textbf{Computational efficiency.}
Figure~\ref{fig:inference_time} shows the per-example inference time on the test set, normalized by the unpruned model. Because our implementation does not exploit sparse kernels, all pruned models incur a runtime overhead of roughly $2$–$2.5\times$ compared to the dense baseline. Within that range, however, the FW-based methods are consistently faster than simple pruning, and FW+momentum is typically the fastest of the three pruning strategies. Thus FW-style pruning does not pay an extra runtime penalty over greedy backward selection in our setup; if anything, it slightly reduces inference time for a given level of sparsity.

\textbf{Key takeaway.}
Across all pre-training budgets, FW+momentum is the only method that simultaneously increases accuracy, decreases test loss, and yields the sparsest models among the algorithms considered. Simple greedy pruning acts more like a noisy perturbation than a principled compression method: it rarely beats the unpruned baseline, barely reduces the parameter count, and is slower at inference time. For practitioners constrained to short pre-training schedules or limited hardware, FW+momentum therefore appears to be the safest default choice among the pruning strategies we tested.

\section{Conclusions}
We studied algorithmic variants of Frank–Wolfe for pruning convolutional neural networks, comparing a simple magnitude-based pruning baseline, an FW-style pruning scheme, and an FW method with momentum. On a CNN trained on MNIST, FW+momentum consistently produced the best trade-off among the methods considered: it achieved higher test accuracy and lower loss than both the dense baseline and the simple pruning method, while also yielding the sparsest models and incurring no additional inference-time overhead in our implementation.

Perhaps most importantly, treating epochs as pre-training budget shows that FW+momentum can be applied after only 1–2 epochs of dense training and still match or surpass the accuracy of fully pre-trained baselines. This suggests that full pre-training of the dense model is unnecessary in this setting, which is especially attractive when compute is limited or rapid model iteration is required. Overall, our results indicate that Frank–Wolfe with momentum is a promising and practical tool for compressing CNNs in resource-constrained environments.

\section{Further Research}

\paragraph{Generalizability and Comparison with Other Pruning Methods }
Despite the positive results attained with our numerical experiments, it is still to be seen if these results can be generalized to other settings. With access to more computational resources, the performance of FW pruning may be evaluated by using optimizers other than Stochastic Gradient Descent (SGD), such as ADAM. Moreover, to further test the generalizability of our FW pruning methods, their performance could be tested on established architectures such as ResNet, AlexNet, or VGG. Likewise, one could also test the generalizability from varying the dataset by using datasets such as Fashion-MNIST, CIFAR-10, or ImageNet. Overall, despite promising results, more computational resources are needed to test the performance of our pruning methods in a general setting. 

Moreover, performing these tests would allow us to compare performance against other pruning methods. To do this, we should suggest using a wide benchmark, such as the one available with ShrinkBench \cite{blalock2020state}, to allow a systematic review of the models' performance.

\paragraph{Structured Pruning}
The pruning paradigm we considered before, which masks certain weights as zeros, is known as unstructured pruning. On the other hand, we have structured pruning, which removes entire units, such as neurons, layers, or filters from the NN/CNN. That is, structured pruning changes the architecture of the network rather than masking weights. Combining the algorithms developed in this paper with structured pruning may lead to even stronger numerical results as model size and complexity could be reduced to a greater extent.

\paragraph{Greedy Forward Selection}\cite{ye2020goodsubnetworks}. Another pruning methodology that could be considered for future research is greedy forward selection. Instead of traditional pruning methods that use backward elimination to remove redundant and Byzantine neurons from the larger network, greedy forward selection begins with an empty model, and sequentially adds neurons from the original network that result in the greatest immediate decrease in the loss function \cite{ye2020goodsubnetworks}. The main advantage of this scheme is that there is no need to prune after pre-training, as a subnetwork is found while training. Hence, this approach may be of particular interest in resource-constrained environments. 

\section{Acknowledgements}
This work originated from a project carried out in Spring 2024 under the supervision of Prof. Anastasios Kyrillidis at Rice University. We thank him for helpful feedback and suggestions.

\clearpage

\bibliographystyle{plain} 
\bibliography{bibliography}


\end{document}